\documentclass{article}

\usepackage{booktabs} % for professional tables

\usepackage{booktabs}       % professional-quality tables
\newcommand{\tabitem}{~~\llap{\textbullet}~~}

\usepackage{comment}
\usepackage{amsmath,amssymb} % define this before the line numbering.
\usepackage{color}

\usepackage{subcaption}
\usepackage{multirow}
\usepackage{makecell}
\usepackage{longtable}
\usepackage{supertabular}

\usepackage[export]{adjustbox}
\usepackage{cellspace}  
\usepackage{enumitem}

\usepackage{hyperref}

\usepackage[accepted]{icml2021}

\icmltitlerunning{Towards Fully Interpretable Deep Neural Networks: Are We There Yet?}

\begin{document}

\twocolumn[
%\icmltitle{Fully Interpretable Convolutional Neural Networks: a Survey}
\icmltitle{Towards Fully Interpretable Deep Neural Networks: Are We There Yet?}



\icmlsetsymbol{equal}{*}

\begin{icmlauthorlist}
\icmlauthor{Sandareka Wickramanayake}{soc}
\icmlauthor{Wynne Hsu}{soc,ids}
\icmlauthor{Mong Li Lee}{soc,ids}
\end{icmlauthorlist}

\icmlaffiliation{soc}{School of Computing, National University of Singapore}
\icmlaffiliation{ids}{Institute of Data Science, National University of Singapore}
\icmlcorrespondingauthor{Sandareka Wickramanayake}{sandaw@comp.nus.edu.sg}

\icmlkeywords{Machine Learning, ICML}

\vskip 0.2in
]



\printAffiliationsAndNotice{} % otherwise use the standard text.

\begin{abstract}
Despite the remarkable performance, Deep Neural Networks (DNNs) behave as black-boxes hindering user trust in Artificial Intelligence (AI) systems. Research on opening black-box DNN can be broadly categorized into post-hoc methods and inherently interpretable DNNs. 
While many surveys have been conducted on post-hoc interpretation methods,  little effort is devoted to  inherently interpretable DNNs.  
This paper provides a review of existing methods to develop DNNs with intrinsic interpretability, with a focus on Convolutional Neural Networks (CNNs). The aim is to understand the current progress towards fully interpretable DNNs that can cater to  different interpretation requirements. 
Finally, we identify gaps in current work and suggest potential research directions.   
\end{abstract}

\section{Introduction}
\label{introduction}

\begin{table*}[htbp]	
	\centering
	\small
	\caption{Summary of pros and cons of post-hoc explanations and interpretable DNNs.}
	\begin{tabular}{p{0.1\textwidth}p{0.38\textwidth}p{0.40\textwidth}}\toprule
		Category &  Pros & Cons  \\ \midrule		
		\multirow{3}{{0.14\textwidth}} {Post-hoc explanations} & 
		{\tabitem No re-training and modifications to the model} & {\tabitem May not reveal the true reasons for the prediction} \\
		& {\tabitem Accuracy is not compromised}  & {\tabitem Require separate modeling efforts} \\
		& {\tabitem May be model-agnostic} &	\\ \midrule
		\multirow{2}{{0.14\textwidth}} {Interpretable DNNs } & \tabitem Explanations  are faithful to the model decision & \tabitem Accuracy may be compromised \\		
		& \tabitem Reasoning process is comprehensible & \tabitem May require domain knowledge \\ 
		\bottomrule
	\end{tabular}
	\label{tab:pros_cons}  	
\end{table*}

Deep Neural Networks (DNNs) have demonstrated impressive performance in a wide range of tasks such as computer vision ~\cite{he2016deep,huang2017densely,liu2016ssd,tan2020efficientdet}, natural language processing~\cite{sutskever2014sequence,peters2018deep}, sentiment analysis~\cite{yang2019xlnet,sachan2019revisiting}, etc. Also, applications of DNNs span many domains, including  robotics, retail, manufacturing, and even safety-critical domains such as healthcare~\cite{Cheng2016a,Che2016,litjens2017}. However, DNNs behave as black-boxes where users can neither understand the decision-making procedure nor the reasons behind their predictions. 
This opaque nature has affected the acceptance and deployment of DNN-based applications.

Research on opening black-box DNNs can be broadly categorized into \textit{post-hoc methods} and \textit{inherently interpretable DNNs}. The objective of post-hoc interpretation methods is to provide insights into already trained models. They aim to uncover either the meaning of the learned features \cite{bau2017,olah2018building} or the rationale behind the model  decisions~\cite{zeiler2014visualizing,Bach2015,Hendricks2016,selvaraju2017grad,park2018multimodal,olah2018building,ghorbani2019towards,yeh2019concept,kim2017tcav,kim2018interpretability,Sandareka2019}. These post-hoc methods can be applied without changing the underline model or retraining it.
 On the other hand, inherently interpretable DNNs require architectural changes or regularizations to provide  intrinsic explanations \cite{li2018deep,melis2018towards,chen2019looks,wickramanayake2101learning}. 
 Table~\ref{tab:pros_cons} gives  a summary of the pros and cons of these two categories.

In this paper, we  review  existing interpretable DNNs, with a focus on CNNs, namely 
DNNs with in-built attention \cite{zhang2014part,zheng2017learning,zhou2018interpretable,zhou2016learning,Pillai2021}, DNNs with prototype-based reasoning \cite{li2018deep,melis2018towards,chen2019looks} and DNNs with feature regularizations \cite{zhang2018interpretable,huang2020interpretable,liang2020training,wickramanayake2101learning}.

\section{DNNs with in-built Attention}

Integrating attention to DNNs
is a common approach for natural language processing~\cite{martins2016softmax,wang2016attention}, domain-specific interpretable DNNs~\cite{chen2019personalized,gao2018interpretable} and fine-grained image classification tasks ~\cite{zheng2017learning,fu2017look,zhuang2020learning}. These models aim to expose the parts of an input the model focuses on for decision making. 

Multi-Attention CNN (MA-CNN) \cite{zheng2017learning}  consists of convolution, channel grouping, and classification sub-networks. Given an input image, MA-CNN feeds the image through a set of convolutional layers and extracts feature maps. Feature maps are then passed to a channel grouping network to obtain multiple attention areas. These attention areas are used to classify the input image. Visualizations of attention maps show that MA-CNN focuses on diverse image areas with strong discrimination ability.

Recurrent-Attention CNN (RA-CNN) ~\cite{fu2017look}  uses an attention proposal sub-network to locate the discriminative areas. It then scales the discriminative areas to examine the details in these areas. Such attention localization at different scales helps to identify class specific regions.

Class Activation Map (CAM)~\cite{zhou2016learning} builds attention-based interpretability into standard CNNs. This method replaces the fully connected (FC) layers with a Global Average Pooling (GAP) layer.  Saliency maps with respect to a class $c$ are generated by multiplying each feature map in the final convolutional layer with its weight corresponding to $c$ and taking the channel-wise summation. The resultant saliency map indicates the image regions the model has focused on to classify the given image to $c$.

CI-GC~\cite{Pillai2021} is a self-supervise learning approach to encourage the model to learn consistent interpretations given an explanation mechanism such as saliency map. A new image is created whereby an input image is placed on a random cell in a $ 2 \times 2$ grid, and other three cells are filled with images selected from classes different from that of the input image. The model is trained to minimize the difference between the saliency map of the newly created image and its targeted interpretation, in addition to maximizing the log-likelihood of the correct class. The targeted interpretation is obtained by placing the saliency map of the input image in the corresponding cell and setting the attribution values in other quadrants to zero.

\section{DNNs with Prototype-based Reasoning}

PrototypeDNN proposed in ~\cite{li2018deep}  trains a convolutional auto-encoder to learn a set of prototypes and a prototype classification network to produce the probability distribution over the classes.  Model decisions are explained by presenting decision-relevant prototypes visualized using the decoder.

~\cite{melis2018towards} propose a  Self-Explainable Neural Network (SENN), which is a generalized version of \cite{li2018deep}. SENN
consists of three networks: (a) a concept encoder to generate interpretable concepts, (b) an input-dependent parameterizer to generate relevance scores of those concepts, and (c) a linear aggregator to combine concepts-relevance score pairs to derive the decision.
 The interpretable concepts can be obtained from expert knowledge or automatically learned with three desiderata:
 \begin{itemize}
 	\vspace*{-0.1in}
 	\item Fidelity -
 	 learned concepts  capture relevant information about the input; 
 	\vspace*{-0.1in}
 	\item Diversity - learned concepts  have minimum overlap;
 	
 	\vspace*{-0.1in}
 	\item Grounding - learned concepts are human understandable. 
 \end{itemize}

ProtoPNet~\cite{chen2019looks} adds a new prototype layer and learns a pre-determined number of prototypes for each class $c$.  The learning objective is designed to ensure that the set of prototypes for class $c$ captures the most relevant concepts for identifying images of that class. In the inference phase,  ProtoPNet compares the latent features of an input image against learned prototypes to evaluate if the input image is from class $c$. The comparison is carried by calculating Euclidean distances between each prototype and all the patches of the latent features that have the same shape as the prototype and inverts the distances into similarity scores. The maximum of these similarity scores indicates how strongly a prototype is present in some patch of the input image. The maximum scores of all the prototypes are multiplied by the weight matrix of class $c$ to give the final score that the input image belongs to $c$.

\begin{figure*}[t!]
	\centering
	%\vspace{0.1in}
	\includegraphics[width=1.0\linewidth]{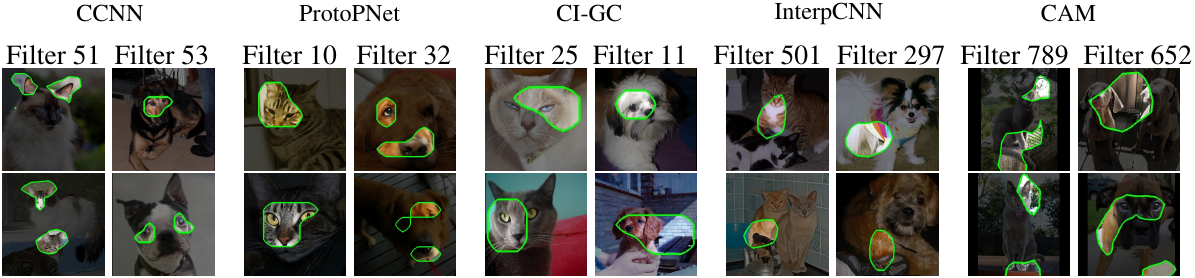}	
	\caption{Visualization of sample conv-filters in CCNN, ProtoPNet, CI-GC, InterpCNN and CAM for the VOC-Part dataset.}	
	\label{fig:comparison_voc}
\end{figure*}

\begin{table*}[htp]
	\caption{Categorization of interpretable DNNs based on  the type of explanations.}
	\centering
	%\ra{1}
	\begin{tabular}{p{2.8cm}p{13.5cm}} \\ \toprule
		% \begin{tabular}{ll} \\ \toprule
		Type of explanation & Methods  \\ \midrule	
		Saliency maps & MA-CNN~\cite{zheng2017learning}, RA-CNN~\cite{fu2017look}, CAM~\cite{zhou2016learning}, Region-CNN~\cite{huang2020interpretable}, CI-GC~\cite{Pillai2021} \\
		Prototypes & PrototypeDNN~\cite{li2018deep}, SENN~\cite{melis2018towards}, ProtoPNet~\cite{chen2019looks} \\
		Word phrases & CCNN~\cite{wickramanayake2101learning} \\	
		\bottomrule
	\end{tabular}
	\label{tab:cat_explanation_type}  
\end{table*}

\section{DNNs with Feature Regularization}

InterpCNN~\cite{zhang2018interpretable} introduces a new loss function to encourage each conv-filter in the higher level convolution layers to be activated only for a specific object part. This loss function maximizes the mutual information between the conv-filter's feature map and a template. A template is the ideal distribution of activations for the feature map given an image. If the given image is from category $c$, a single location of the feature map should be activated. Otherwise, the feature map should remain inactivated.  The training uses image-level labels without the need for object part-level annotations. The  filter loss pushes the feature map of a conv-filer towards representing a specific object part during the learning process.

Region-CNN~\cite{huang2020interpretable}  is an interpretable classifier specific for fine-grained recognition. It learns to group pixels into meaningful object part regions by enforcing a prior distribution for the occurrence of each part.

CCNN~\cite{wickramanayake2101learning} adds an additional concept layer to a CNN-based architecture to guide the learning of the associations between visual features and word phrases extracted from image descriptions.
%This allows CCNN to explain its decisions in terms of word phrases corresponding to different visual concepts. 
The training objective function  takes into consideration concept uniqueness and  mapping consistency. The former 
encourages each  learned concept to  correspond to only one  word phrase, while the latter  aims to preserve the distance between the learned concept and its corresponding word phrase in a joint embedding space. Together with classification accuracy loss, this training objective ensures that CCNN is both accurate and interpretable. CCNN employs a GAP layer to reduce the dimensionality of the concept layer's outputs before feeding to a fully connected layer. Hence, CCNN classification decisions can be expressed as a weighted sum of the learned concepts. In other words, CCNN can explain its decisions in terms of  word phrases and their corresponding contributions.

\section{Discussion}

We observe that most of the interpretable DNNs described in the previous section make decisions based on a linear aggregation of the 
learned interpretable features. The difference lies in 
how the interpretable features are defined and how such features are learned. For example, InterpCNN uses disentangled concepts, ProtoPNet and SENN use prototypes and CCNN uses features that  correspond to word phrases. 
  Figure~\ref{fig:comparison_voc}  shows the features learned by CCNN, ProtoPNet, CI-GC, InterpCNN and CAM taken from ~\cite{wickramanayake2101learning}.
  We see that some of these features   may not correspond to concepts consistent with human perception. For instance, although InterpCNN ~\cite{zhang2018interpretable} strives to learn features that correspond to  high-level concepts, the learned features  cover only part of the region  when the concept spans multiple non-contiguous regions in the image, % such as the eyes of a dog.
  e.g., when an image of two cats is given, the filter has activated for the chest of only one cat.

Another observation  is that these interpretable DNNs differ in the type of explanations they provide, from saliency maps to prototypes to word phrases (see Table~\ref{tab:cat_explanation_type}).
We see that CCNN is the only one offering explanations in the form of word phrases. 
Further, many of these interpretable models can only explain an individual data point's prediction, whereas ProtoPNet and CCNN  can  additionally provide the concepts important for predicting a certain class.
For example, ProtoPNet can show the prototypes that contribute to the highest number of correct classifications of a class, while  CCNN can give the word phrases corresponding to the set of visual features that is activated by the largest number of images in the class.

We also notice inconsistencies among the evaluation studies in existing work. While interpretable DNNs should be evaluated for both predictive performance and interpretability, some of the current work have neglected either of aspects, e.g., SENN is not assessed for classification accuracy whereas ProtoPNet is not quantitatively evaluated for its interpretability. Further, although it is intuitive that explanations intrinsic to the model are more faithful than post-hoc explanations, systematic evaluations should be provided to convince the user. Currently, only ~\cite{melis2018towards} conduct experiments to evaluate the faithfulness of generated explanations to the model decision.

\section{Conclusion and Future Research}

Regardless of how interpretable DNNs generate their  explanations, we believe that there is a need to cater to 
different users who may want to understand the DNNs from different perspectives.  
For instance, one might be interested in understanding the rationale behind a specific decision of the model, while another wants to know what concepts are used by the model to differentiate a given class. Another user might be keen to understand what changes in the input would change the model prediction to a pre-defined output. 
Ideally, a fully interpretable system should be able to provide all these
explanations, which
 none of the existing interpretable DNNs are able to. Investigating ways to provide  explanations from multiple perspectives is a promising research direction.

Another direction is to have a unified criterion to evaluate interpretability of learned features. 
Existing work have introduced various metrics such as 
part interpretability and location instability~\cite{zhang2018interpretable},  homogeneity and  single-ness~\cite{wickramanayake2101learning}.  The first three metrics indicate if a conv-filter is activated for the same concept across multiple images, while the last metric single-ness indicates if a  conv-filter is activated for a unique concept in a given image. We believe that the notion of \textit{an interpretable feature} being ill-defined has led to numerous metrics.
Given that feature interpretability is an essence of a fully interpretable DNN, having a unified objective metric to evaluate feature interpretability would enable researchers to benchmark different inherently interpretable DNNs.

Finally,  developing interpretable DNNs that provide semantic explanations is another promising research direction.
We see that most explanations provided by current interpretable DNNs are visual explanations, either in the form of saliency maps or prototypes. However, visual explanations may be ambiguous and the ability to explain DNN decisions using semantic concepts are more useful \cite{Sandareka2019,chen2019explaining}. The work in \cite{wickramanayake2101learning} has taken the first  step in this direction by using image descriptions to learn visual concepts that are consistent with human-understandable concepts. \\

\noindent\textbf{Acknowledgement.} This research is supported by the National
Research Foundation Singapore under its AI Singapore
Programme (AISG-GC-2019-001, AISG-RP-2018-008).

\bibliographystyle{icml2021}
\bibliography{references}

\begin{thebibliography}{42}
\providecommand{\natexlab}[1]{#1}
\providecommand{\url}[1]{\texttt{#1}}
\expandafter\ifx\csname urlstyle\endcsname\relax
  \providecommand{\doi}[1]{doi: #1}\else
  \providecommand{\doi}{doi: \begingroup \urlstyle{rm}\Url}\fi

\bibitem[Bach et~al.(2015)Bach, Binder, Montavon, Klauschen, M{\"{u}}ller, and
  Samek]{Bach2015}
Bach, S., Binder, A., Montavon, G., Klauschen, F., M{\"{u}}ller, K.~R., and
  Samek, W.
\newblock {On pixel-wise explanations for non-linear classifier decisions by
  layer-wise relevance propagation}.
\newblock \emph{PLoS ONE}, 2015.

\bibitem[Bau et~al.(2017)Bau, Zhou, Khosla, Oliva, and Torralba]{bau2017}
Bau, D., Zhou, B., Khosla, A., Oliva, A., and Torralba, A.
\newblock Network dissection: Quantifying interpretability of deep visual
  representations.
\newblock In \emph{CVPR}, 2017.

\bibitem[Che et~al.(2016)Che, Purushotham, Khemani, and Liu]{Che2016}
Che, Z., Purushotham, S., Khemani, R., and Liu, Y.
\newblock {Interpretable Deep Models for ICU Outcome Prediction.}
\newblock \emph{AMIA Annual Symposium Proceedings}, 2016.

\bibitem[Chen et~al.(2019{\natexlab{a}})Chen, Li, Tao, Barnett, Rudin, and
  Su]{chen2019looks}
Chen, C., Li, O., Tao, D., Barnett, A., Rudin, C., and Su, J.~K.
\newblock This looks like that: deep learning for interpretable image
  recognition.
\newblock In \emph{NeurIPS}, 2019{\natexlab{a}}.

\bibitem[Chen et~al.(2019{\natexlab{b}})Chen, Chen, Ren, Huang, and
  Zhang]{chen2019explaining}
Chen, R., Chen, H., Ren, J., Huang, G., and Zhang, Q.
\newblock Explaining neural networks semantically and quantitatively.
\newblock In \emph{ICCV}, 2019{\natexlab{b}}.

\bibitem[Chen et~al.(2019{\natexlab{c}})Chen, Chen, Xu, Zhang, Cao, Qin, and
  Zha]{chen2019personalized}
Chen, X., Chen, H., Xu, H., Zhang, Y., Cao, Y., Qin, Z., and Zha, H.
\newblock Personalized fashion recommendation with visual explanations based on
  multimodal attention network: Towards visually explainable recommendation.
\newblock In \emph{SIGIR}, 2019{\natexlab{c}}.

\bibitem[Cheng et~al.(2016)Cheng, Wang, Zhang, and Hu]{Cheng2016a}
Cheng, Y., Wang, F., Zhang, P., and Hu, J.
\newblock {Risk Prediction with Electronic Health Records: A Deep Learning
  Approach}.
\newblock In \emph{SIAM International Conference on Data Mining}, pp.\
  432--440, 2016.

\bibitem[Fu et~al.(2017)Fu, Zheng, and Mei]{fu2017look}
Fu, J., Zheng, H., and Mei, T.
\newblock Look closer to see better: Recurrent attention convolutional neural
  network for fine-grained image recognition.
\newblock In \emph{CVPR}, 2017.

\bibitem[Gao et~al.(2018)Gao, Fokoue, Luo, Iyengar, Dey, and
  Zhang]{gao2018interpretable}
Gao, K.~Y., Fokoue, A., Luo, H., Iyengar, A., Dey, S., and Zhang, P.
\newblock Interpretable drug target prediction using deep neural
  representation.
\newblock In \emph{IJCAI}, 2018.

\bibitem[Ghorbani et~al.(2019)Ghorbani, Wexler, Zou, and
  Kim]{ghorbani2019towards}
Ghorbani, A., Wexler, J., Zou, J.~Y., and Kim, B.
\newblock Towards automatic concept-based explanations.
\newblock In \emph{NeurIPS}, 2019.

\bibitem[He et~al.(2016)He, Zhang, Ren, and Sun]{he2016deep}
He, K., Zhang, X., Ren, S., and Sun, J.
\newblock Deep residual learning for image recognition.
\newblock In \emph{CVPR}, 2016.

\bibitem[Hendricks et~al.(2016)Hendricks, Akata, Rohrbach, Donahue, Schiele,
  and Darrell]{Hendricks2016}
Hendricks, L.~A., Akata, Z., Rohrbach, M., Donahue, J., Schiele, B., and
  Darrell, T.
\newblock {Generating visual explanations}.
\newblock In \emph{ECCV}, 2016.

\bibitem[Huang et~al.(2017)Huang, Liu, Van Der~Maaten, and
  Weinberger]{huang2017densely}
Huang, G., Liu, Z., Van Der~Maaten, L., and Weinberger, K.~Q.
\newblock Densely connected convolutional networks.
\newblock In \emph{CVPR}, 2017.

\bibitem[Huang \& Li(2020)Huang and Li]{huang2020interpretable}
Huang, Z. and Li, Y.
\newblock Interpretable and accurate fine-grained recognition via region
  grouping.
\newblock In \emph{CVPR}, 2020.

\bibitem[Kim et~al.(2017)Kim, Gilmer, Viegas, Erlingsson, and
  Wattenberg]{kim2017tcav}
Kim, B., Gilmer, J., Viegas, F., Erlingsson, U., and Wattenberg, M.
\newblock Tcav: Relative concept importance testing with linear concept
  activation vectors.
\newblock \emph{arXiv}, 2017.

\bibitem[Kim et~al.(2018)Kim, Wattenberg, Gilmer, Cai, Wexler, Viegas,
  et~al.]{kim2018interpretability}
Kim, B., Wattenberg, M., Gilmer, J., Cai, C., Wexler, J., Viegas, F., et~al.
\newblock Interpretability beyond feature attribution: Quantitative testing
  with concept activation vectors (tcav).
\newblock In \emph{ICML}, 2018.

\bibitem[Li et~al.(2018)Li, Liu, Chen, and Rudin]{li2018deep}
Li, O., Liu, H., Chen, C., and Rudin, C.
\newblock Deep learning for case-based reasoning through prototypes: A neural
  network that explains its predictions.
\newblock In \emph{AAAI}, 2018.

\bibitem[Liang et~al.(2020)Liang, Ouyang, Zeng, Su, He, Xia, Zhu, and
  Zhang]{liang2020training}
Liang, H., Ouyang, Z., Zeng, Y., Su, H., He, Z., Xia, S.-T., Zhu, J., and
  Zhang, B.
\newblock Training interpretable convolutional neural networks by
  differentiating class-specific filters.
\newblock In \emph{ECCV}, 2020.

\bibitem[Litjens et~al.(2017)Litjens, Kooi, Bejnordi, Arindra, Setio, Ciompi,
  Ghafoorian, {Van Der Laak}, {Van Ginneken}, and S{\'{a}}nchez]{litjens2017}
Litjens, G., Kooi, T., Bejnordi, B.~E., Arindra, A., Setio, A., Ciompi, F.,
  Ghafoorian, M., {Van Der Laak}, J. A. W.~M., {Van Ginneken}, B., and
  S{\'{a}}nchez, C.~I.
\newblock {A Survey on Deep Learning in Medical Image Analysis}.
\newblock \emph{Medical image analysis}, 2017.

\bibitem[Liu et~al.(2016)Liu, Anguelov, Erhan, Szegedy, Reed, Fu, and
  Berg]{liu2016ssd}
Liu, W., Anguelov, D., Erhan, D., Szegedy, C., Reed, S., Fu, C.-Y., and Berg,
  A.~C.
\newblock Ssd: Single shot multibox detector.
\newblock In \emph{ECCV}, 2016.

\bibitem[Martins \& Astudillo(2016)Martins and Astudillo]{martins2016softmax}
Martins, A. and Astudillo, R.
\newblock From softmax to sparsemax: A sparse model of attention and
  multi-label classification.
\newblock In \emph{ICML}, 2016.

\bibitem[Melis \& Jaakkola(2018)Melis and Jaakkola]{melis2018towards}
Melis, D.~A. and Jaakkola, T.
\newblock Towards robust interpretability with self-explaining neural networks.
\newblock In \emph{NeurIPS}, 2018.

\bibitem[Olah et~al.(2018)Olah, Satyanarayan, Johnson, Carter, Schubert, Ye,
  and Mordvintsev]{olah2018building}
Olah, C., Satyanarayan, A., Johnson, I., Carter, S., Schubert, L., Ye, K., and
  Mordvintsev, A.
\newblock The building blocks of interpretability.
\newblock \emph{Distill}, 2018.

\bibitem[Park et~al.(2018)Park, Hendricks, Akata, Rohrbach, Schiele, Darrell,
  and Rohrbach]{park2018multimodal}
Park, D.~H., Hendricks, L.~A., Akata, Z., Rohrbach, A., Schiele, B., Darrell,
  T., and Rohrbach, M.
\newblock Multimodal explanations: Justifying decisions and pointing to the
  evidence.
\newblock In \emph{CVPR}, 2018.

\bibitem[Peters et~al.(2018)Peters, Neumann, Iyyer, Gardner, Clark, Lee, and
  Zettlemoyer]{peters2018deep}
Peters, M.~E., Neumann, M., Iyyer, M., Gardner, M., Clark, C., Lee, K., and
  Zettlemoyer, L.
\newblock Deep contextualized word representations.
\newblock In \emph{ACL}, 2018.

\bibitem[Pillai \& Pirsiavash(2021)Pillai and Pirsiavash]{Pillai2021}
Pillai, V. and Pirsiavash, H.
\newblock Explainable models with consistent interpretations.
\newblock In \emph{AAAI}, 2021.

\bibitem[Sachan et~al.(2019)Sachan, Zaheer, and
  Salakhutdinov]{sachan2019revisiting}
Sachan, D.~S., Zaheer, M., and Salakhutdinov, R.
\newblock Revisiting lstm networks for semi-supervised text classification via
  mixed objective function.
\newblock In \emph{AAAI}, 2019.

\bibitem[Selvaraju et~al.(2017)Selvaraju, Cogswell, Das, Vedantam, Parikh, and
  Batra]{selvaraju2017grad}
Selvaraju, R.~R., Cogswell, M., Das, A., Vedantam, R., Parikh, D., and Batra,
  D.
\newblock Grad-cam: Visual explanations from deep networks via gradient-based
  localization.
\newblock In \emph{ICCV}, 2017.

\bibitem[Sutskever et~al.(2014)Sutskever, Vinyals, and
  Le]{sutskever2014sequence}
Sutskever, I., Vinyals, O., and Le, Q.~V.
\newblock Sequence to sequence learning with neural networks.
\newblock In \emph{NeurIPS}, 2014.

\bibitem[Tan et~al.(2020)Tan, Pang, and Le]{tan2020efficientdet}
Tan, M., Pang, R., and Le, Q.~V.
\newblock Efficientdet: Scalable and efficient object detection.
\newblock In \emph{CVPR}, 2020.

\bibitem[Wang et~al.(2016)Wang, Huang, Zhu, and Zhao]{wang2016attention}
Wang, Y., Huang, M., Zhu, X., and Zhao, L.
\newblock Attention-based lstm for aspect-level sentiment classification.
\newblock In \emph{EMNLP}, 2016.

\bibitem[Wickramanayake et~al.(2019)Wickramanayake, Hsu, and
  Lee]{Sandareka2019}
Wickramanayake, S., Hsu, W., and Lee, M.~L.
\newblock Flex: Faithful linguistic explanations for neural net based model
  decisions.
\newblock In \emph{AAAI}, 2019.

\bibitem[Wickramanayake et~al.(2021)Wickramanayake, Hsu, and
  Lee]{wickramanayake2101learning}
Wickramanayake, S., Hsu, W., and Lee, M.~L.
\newblock Comprehensible convolutional neural networks via guided concept
  learning.
\newblock \emph{arXiv preprint arXiv:2101.03919}, 2021.

\bibitem[Yang et~al.(2019)Yang, Dai, Yang, Carbonell, Salakhutdinov, and
  Le]{yang2019xlnet}
Yang, Z., Dai, Z., Yang, Y., Carbonell, J., Salakhutdinov, R., and Le, Q.~V.
\newblock Xlnet: Generalized autoregressive pretraining for language
  understanding.
\newblock In \emph{NeurIPS}, 2019.

\bibitem[Yeh et~al.(2019)Yeh, Kim, Arik, Li, Ravikumar, and
  Pfister]{yeh2019concept}
Yeh, C.-K., Kim, B., Arik, S.~O., Li, C.-L., Ravikumar, P., and Pfister, T.
\newblock On concept-based explanations in deep neural networks.
\newblock \emph{arXiv preprint}, 2019.

\bibitem[Zeiler \& Fergus(2014)Zeiler and Fergus]{zeiler2014visualizing}
Zeiler, M.~D. and Fergus, R.
\newblock Visualizing and understanding convolutional networks.
\newblock In \emph{ECCV}, pp.\  818--833, 2014.

\bibitem[Zhang et~al.(2014)Zhang, Donahue, Girshick, and
  Darrell]{zhang2014part}
Zhang, N., Donahue, J., Girshick, R., and Darrell, T.
\newblock Part-based r-cnns for fine-grained category detection.
\newblock In \emph{ECCV}, 2014.

\bibitem[Zhang et~al.(2018)Zhang, Nian~Wu, and Zhu]{zhang2018interpretable}
Zhang, Q., Nian~Wu, Y., and Zhu, S.-C.
\newblock Interpretable convolutional neural networks.
\newblock In \emph{CVPR}, 2018.

\bibitem[Zheng et~al.(2017)Zheng, Fu, Mei, and Luo]{zheng2017learning}
Zheng, H., Fu, J., Mei, T., and Luo, J.
\newblock Learning multi-attention convolutional neural network for
  fine-grained image recognition.
\newblock In \emph{ICCV}, 2017.

\bibitem[Zhou et~al.(2016)Zhou, Khosla, Lapedriza, Oliva, and
  Torralba]{zhou2016learning}
Zhou, B., Khosla, A., Lapedriza, A., Oliva, A., and Torralba, A.
\newblock Learning deep features for discriminative localization.
\newblock In \emph{CVPR}, 2016.

\bibitem[Zhou et~al.(2018)Zhou, Sun, Bau, and Torralba]{zhou2018interpretable}
Zhou, B., Sun, Y., Bau, D., and Torralba, A.
\newblock Interpretable basis decomposition for visual explanation.
\newblock In \emph{ECCV}, 2018.

\bibitem[Zhuang et~al.(2020)Zhuang, Wang, and Qiao]{zhuang2020learning}
Zhuang, P., Wang, Y., and Qiao, Y.
\newblock Learning attentive pairwise interaction for fine-grained
  classification.
\newblock In \emph{AAAI}, 2020.

\end{thebibliography}

\end{document}